\documentclass{svjour3}                     
\smartqed  
\usepackage{bbding}
\usepackage[misc]{ifsym}
\usepackage{graphicx}
\usepackage{float}
 \usepackage{latexsym,bm,amsmath,amssymb} 
\usepackage{color}
\usepackage[ruled]{algorithm2e}
\usepackage{cite}
\usepackage{caption}
\usepackage{subfig}
\usepackage{makecell}

\newtheorem{pro}{Proposition}
\newtheorem{mydef}{Definition}
\begin{document}

\title{Volume Preserving Image Segmentation with Entropic Regularization Optimal Transport and Its Applications in Deep Learning\thanks{Liu and Huang were partly 
supported by The National Key Research and Development Program of China (No. 2017YFA0604903). Liu was also partly supported by The National Natural Science Foundation of China (No. 11871035).}
}


\author{Haifeng Li \textsuperscript{1}\and Jun Liu \textsuperscript{1,\Letter}\and
        Li Cui \textsuperscript{1,\Letter} \and Haiyang Huang \textsuperscript{1} \and Xue-cheng Tai \textsuperscript{2}
}


\institute{\at {1.}
              Laboratory of Mathematics and Complex Systems (Ministry of Education of China), School of Mathematical Sciences, Beijing Normal University, Beijing, 100875, People's Republic of China. \\                   
                      \at   {2.} Department of Mathematics, Hong Kong Baptist University, Kowloon Tong, Hong Kong.\\
                      \at {\Letter} Co-corresponding authors, \email{jliu@bnu.edu.cn,~licui@bnu.edu.cn}.
}


\maketitle

\begin{abstract}
Image segmentation with a volume constraint is an important prior for many real applications. 
In this work, we present a novel volume preserving image segmentation algorithm, which is based on the framework of entropic regularized optimal transport theory. The classical Total Variation (TV) regularizer and volume preserving are integrated into a regularized optimal transport model, and the volume and classification constraints can be regarded as two measures preserving constraints in the optimal transport problem. By studying the dual problem, we develop a simple and efficient dual algorithm for our model. Moreover, to be different from many variational based image segmentation algorithms, the proposed algorithm can be directly unrolled to a new Volume Preserving and TV regularized softmax (VPTV-softmax) layer for semantic segmentation in the popular Deep Convolution Neural Network (DCNN). The experiment results show that our proposed model is very competitive and can improve the performance of many semantic segmentation nets such as the popular U-net and DeepLabv3+.
\keywords{Image segmentation \and volume preserving  \and optimal transport  \and entropic regularization  \and TV regularization \and U-net \and DeepLabv3+}
\end{abstract}

\section{Introduction}
Image segmentation is to part an image into several non-overlapping regions according to different similarities. It is a fundamental technique in computer vision.  A large number of algorithms have been developed for this problem. In which
handcraft model and learning based methods are two major tools.
For handcraft model, the image segmentation priors such as shapes, smoothness, volume constraints are manually designed, while these features can be learned in learning based methods. The variational and statistics methods may be two popular handcraft techniques. 
In the variational framework, segmentation models usually contain two terms. One of them is called fidelity term, which measures the similarity of regions. The other  is the regularization term, which smoothes the segmentation boundaries, and avoids the influence of noises. The representative variational models include the Mumford-Shan\cite{mumford1989optimal}, Chan-Vese \cite{chan2001active}, geodesic active contours models \cite{caselles1997geodesic} and so on. Statistics based image segmentation approaches also have been widely developed in literatures. In this method, usually, pixels in an image can be modeled as many realizations of a finite mixture probability distributions such as Gaussian Mixture Model (GMM) \cite{permuter2003gaussian, dempster1977maximum}, Markov Random Field (MRF) model and Bayesian method \cite{geman1984stochastic, besag1989digital} \emph{etc.}. Statistics based methods are more natural and good at dealing with the big data. However, these methods cannot detect an accurate object boundary if the images contain noises. On the other hand, the variational methods can incorporate geometric information more flexibly. Recently, it has been shown that the combination of statistics and variational based approaches can give very good results. For example, in \cite{liu2012expectation}, the authors proposed a unified variational method to bring together the Expectation Maximum (EM) algorithm and Total Variation (TV) regularization and to take advantages from both approaches. 
Inspired by this, we develop a new image segmentation scheme which also  integrates the both advantages. Meanwhile, we will show that our proposed method is a volume constrained version of EMTV.

Volume preserving segmentation means that each phase must preserve a certain volume constraint, in which volume is applied as a prior. This method can greatly improve the segmentation accuracy and has many real applications. For example, in some particular medical image segmentation problems, we can obtain the volume prior by experience or other means, and the segmentation results can be improved by volume prior. On the other hand, for some poor quality images, such as noisy image, edge blurred image, shaded image and so on, volume preserving will play a crucial role. Jacobs \emph{et al.} in \cite{jacobs2018auction} proposed a volume constrained MBO scheme, called auction dynamics scheme, which adding volume constraint conditions to the threshold dynamics segmentation. In \cite{jacobs2018auction}, the authors numerically showed that volume preserving can increase the segmentation accuracy especially when the training set is extremely small. More notably, this advantage persists even one can only estimate very rough volume bound, which again illustrates the benefits of the volume preserving. However, the MBO scheme is an approximated length term of classification boundaries, and its accuracy for segmentation  is lower than TV's. What is more, MBO can not be directly replaced by TV  in their method since TV is non-smooth and convex (convexity lead to the linearization fails in this method), and it would lead to the algorithm fail to converge. We will show the detailed reasons in Section 2. In our proposed model, we will solve these problems by introducing the dual formulation of TV and entropy regularizer. Another work closely related to our method is a global minimization of continuous multiphase partitioning method \cite{Bae2011}, which can be interpreted as using the dual formulation of TV and entropy regularizer in Potts model \cite{Potts1952}, and this will be detailed in Section 2.

Another novelty of this work is to consider the proposed model under the framework of optimal transportation and regard the volume and classification constraints as measure preserving appeared in optimal transport. To the best of our knowledge, this has been first explored in \cite{ni2009local} for comparing local 1D features. Then \cite{peyre2012wasserstein} proposed an active contours approach, using approximate Wasserstein distance for comparing global multi-dimensional features of the interested region. However, these methods are sensitive to the initial contour because of its nonconvexity. In \cite{yildizouglu2013convex}, a convex formulation for two-phase image segmentation was proposed, then was extended to a regularized \cite{Cuturi2013Sinkhorn} optimal transport distance in \cite{rabin2015convex}, where the fixed exemplar histograms define a prior on the statistical features of the two regions in competition. Then \cite{rabin2015convex} was extended to multi-phase image segmentation in \cite{papadakis2017convex}. It is well-know that the Wasserstein distance from optimal transport is time-consuming in real implementation since this distance is described by an intractable optimal problem. In \cite{Cuturi2013Sinkhorn}, the authors developed an entropic regularization method to approximate the solution of optimal transport problem, which can be efficiently solved by the Sinkhorn algorithm \cite{Cuturi2013Sinkhorn}. Inspired by this work, we will adopt entropy regularization in our method.

Instead of using histograms in the above mentioned methods, in this paper, we consider the volume and classification constraints as measure preserving appeared in optimal transport. Thus the problem of volume constrained image segmentation can be regarded as a transportation problem between the volume preserving and pixel classification constraint. In order to integrate the superiority of TV, we adopt both entropic and TV regularization in the optimal transport. It must be particularly pointed out that the standard dual method for TV regularized optimal transport 
is not stable without entropic regularization. To efficiently solve the proposed model, we develop a converged dual algorithm.

Recently, the Deep Convolution Neural Network (DCNN)
based learning method is very successful in image segmentation. For example, U-net\cite{Ronneberger2015}, Seg-net \cite{Badrinarayanan2015} and DeepLabv3+ \cite{chen2018encoder} are successful for image semantic segmentation.
In these mentioned networks, the key classification function adopted in the last layer usually is  the softmax
activation function. However, the classical  softmax does not
contain any spatial and volume prior. In this paper, we can show that the proposed algorithm can be simply unrolled 
as a Volume Preserving and TV regularized softmax (VPTV-sfotmax). The new VPTV-softmax can integrate some handcraft prior into learning method, which can improve the 
performance of the networks.

The main contributions of the paper include:
\begin{itemize}
\item TV regularization and volume preserving are integrated into an entropy regularization optimal transport for image segmentation. 
\item An efficient and stable dual algorithm for the volume preserving image segmentation is developed.
\item A novel volume preserving and TV regularized softmax
layer for deep learning based semantic segmentation is proposed.
It can combine the superiority of the machine learning and handcraft models.  We 
give a general DCNN framework to consider the local prior such as smooth boundaries, volume prior appeared in many applications. Moreover, the new network layers are derived from a continuous smooth dual algorithm, which 
has a mathematical theory. 
\item For the practice applications in compute vision, the proposed method can greatly improve the performance of many popular segmentation nets such as U-net and DeepLabv3+ no matter what the network architectures are.
\end{itemize}

The remainder of this paper is organized as follows: in section 2, the related works are reviewed; section 3 contains
the proposed volume constrained image segmentation model
including model and algorithm; in section 4, we show the connections of many segmentation algorithms and the proposed approach; section 5 includes the extension of the proposed algorithm in deep learning;
in section 6, we show some numerical experiments to demonstrate the effectiveness of our model; finally, a brief conclusion is drawn in section 7.

\section{The related works}
In this section, we will review the related classical image segmentation works.
\subsection{Potts model}
A classical variational based image segmentation model 
is the Potts model \cite{Potts1952}. Let $h:\Omega\subset\mathbb{R}^2\rightarrow\mathbb{R}^d$ be an image ($d=1$ for gray and $d=3$ for color images), and the discrete set $\Omega=\{x_j\}_{j=1}^J$, where $J$ is the number of pixels of image $h$, and $I$ the number of phases, then
the relaxed version of Potts model can be written as 
\begin{equation}\label{Potts}
\mathop{\mathrm{min}}\limits_{\boldsymbol{u}\in\mathbb{U}}\sum\limits_{i=1}^I\sum\limits_{j=1}^Jc_i(x_j)u_i(x_j)+\lambda \sum\limits_{i=1}^I\sum\limits_{j=1}^J ||\nabla u_i(x_j)||,
\end{equation}
where $c_i(x_j)$ is a similarity term to measure the intensity of pixel $h(x_j)$ belongs to $i$-th phase. For example,
$c_i(x_j)=||(h(x_j)-m_i)||^2$ ($m_i$ is the mean of image intensity in the $i$-th region).
 While the second term is the well-know TV \cite{Rudin1992}, which is strictly equal to the total length of 
boundaries of the segmented regions if $u_i$ is the indicator function of the $i$-th region $\Omega_i$.
$\lambda$ is a regularization parameter which control the balance of these two terms.
 The set
$$ \mathbb{U}=\left\{\bm u\in[0,1]^{I}:~\sum\limits_{i=1}^I u_i(x_j)=1, \forall x_j\in \Omega.\right\}$$ 
is a simplex which forms a segmentation condition. 


\subsection{Threshold dynamics volume constrained image segmentation}

In \cite{jacobs2018auction}, Jacobs \emph{et al.} developed a threshold dynamics method (or call MBO scheme \cite{merriman1992diffusion}) for image segmentation in the presence of volume constraint of phases. Their model can be written as the following optimization problem 
\begin{equation}{\label{ThreshD}}
\mathop{\mathrm{min}}\limits_{\boldsymbol{u}\in\mathbb{U}(\bm V)}\sum\limits_{i=1}^I\sum\limits_{j=1}^Jc_i(x_j)u_i(x_j)+\lambda \sum\limits_{i=1}^I\sum\limits_{j=1}^J u_i(x_j)(k*(1-u_i))(x_j),
\end{equation}
The symbol ``*'' stands for convolution and $k$ is a kernel function (usually can be Gaussian kernel) and 
 $$ \mathbb{U}(\bm V)=\left\{\bm u\geqslant 0:~\sum\limits_{i=1}^I u_i(x_j)=1, \forall x_j\in \Omega, ~\sum\limits_{j=1}^Ju_i(x_j)=V_i=|\Omega_i|, \forall  i=1,2,\cdots,I.\right\}.$$
The last constraint condition in $\mathbb{U}(\bm V)$ is a volume constraint, and $\bm V=(V_1,V_2,\cdots,V_I)$ means that each phase $i$ should be occupied with $V_i$ pixels. Here $V_i$ is a known quantity and the sum of $V_i$ equals to the total number of pixels of image $h$, i.e. $\sum_{i=1}^I V_i=|\Omega|=J$.

Compared with Potts model, the second term in \eqref{ThreshD} is not TV but an approximated length penalty. What is more, the second condition in $\mathbb{U}(\bm V)$ is a volume constraint, and it can greatly improve the performance of the segmentation algorithm if the volume prior is given. 

The difficulty of this minimization problem \eqref{ThreshD} is that the regularization term, let us denote as $\mathcal{R}(\bm u)$, is concave. Fortunately, it can be linearized as 
$$\hat{\mathcal{R}}(\bm u;\bm u^{t})= \sum\limits_{i=1}^I\sum\limits_{j=1}^J u_i(x_j)(k*(1-u^{t}_i))(x_j),$$
where $t$ is the iteration number. Such a linearization was earlier adopted and studied in \cite{Liu2011} from constraint optimization. The similar idea was also be studied as iterative thresholding method in \cite{Dong2017} recently.
With this linearization, the problem \eqref{ThreshD} becomes a linear problem in each iteration, and it can be shown in \cite{jacobs2018auction} that this problem is equivalent to an optimal transport problem. In \cite{jacobs2018auction}, to keep the two constraints of $\mathbb{U}(\bm V)$, a discrete auction algorithm is adopted and giving a binary output of $\bm u$. This algorithm is efficient, however, the image segmentation accuracy is not good as TV due to the approximation of length term in the cost functional.

\subsection{EMTV model}
The EMTV model  \cite{liu2012expectation} combines Expectation Maximum (EM) algorithm with TV regularization to implement image segmentation. It integrates the advantages of both the EM algorithm and the TV-based method. Compared with the classical EM algorithm, EMTV has a geometrical constraint, which makes this algorithm robust for noise. On the other hand, compared with the level set method, the statistical information is taken into the model and it is more suitable for natural images. Furthermore, this algorithm can conveniently handle multi-clusters and does not need any extra method to reinitialize the level set functions. 

Theoretically, the fundamental principles of the EM algorithm and the $\mathrm{TVL}^1$  model are very different. One way to overcome the difficulty is to reinterpret the EM algorithm using a constraint optimization framework \cite{liu2011simultaneous, liu2013weighted}.

The EM types image segmentation are based on the Maximum Likelihood Estimation (MLE) of a Gaussian Mixture Model (GMM) model.
In continuous case, the negative log-likelihood function of GMM is given by
\begin{equation}
\mathcal{L}(\Theta)=-\int_\Omega\log\sum\limits_{i=1}^I\alpha_ip_i(h(x);\boldsymbol{ \mu}_i,\boldsymbol{ \sigma}_i)dx,
\end{equation}
where  $\Theta=(\alpha_1,...,\alpha_I,\boldsymbol{ \mu}_1,...,\boldsymbol{ \mu}_I,\boldsymbol{ \sigma}_1,...,\boldsymbol{ \sigma}_I)$ is a parameter vector. This function contains a log-sum functional, which makes it difficult to be directly optimized because the logarithm and summation operations are in general non-commutative. In \cite{liu2011simultaneous}, to overcome this difficulty, the authors introduced a constraint optimization framework with a convex relaxation method to optimize the log-sum type functional. In this way, they showed that the constraint optimization method is essentially equivalent to the EM algorithm. The key idea is the following proposition. 

\begin{pro}[Commutativity of Log-sum operation \cite{liu2013weighted, teboulle2007unified}]\label{pro4}
Given two functions $\alpha_i(x)>0$, $p_i(x)>0$, one can get
$$-\log\sum\limits_{i=1}^I\alpha_i(x)p_i(x)=
\mathop{\mathrm{min}}\limits_{\boldsymbol{u(x)}\in\mathbb{U}}\left\{-\sum\limits_{i=1}^I\log[\alpha_i(x)p_i(x)]u_i(x)+\sum\limits_{i=1}^Iu_i(x)\log u_i(x)\right\},$$
where $\boldsymbol{u}(x)=(u_1(x),...,u_I(x))$ and $\mathbb{U}=\left\{\boldsymbol{u}(x):0\leqslant u_i(x)\leqslant 1, \sum\limits_{i=1}^Iu_i(x)=1\right\}$.
\end{pro}
Using this proposition, we can obtain a new functional $\mathcal{H}(\boldsymbol{u},\Theta)$ with an additional variable $\boldsymbol{u}$ 
\begin{equation}   
\mathcal{H}(\boldsymbol{u},\Theta)=-\int_\Omega\sum\limits_{i=1}^I\log[\alpha_ip_i(I(x);\boldsymbol{ \mu}_i,\boldsymbol{ \sigma}_i)]u_i(x)dx+\int_\Omega\sum\limits_{i=1}^Iu_i(x)\log u_i(x)dx.
\label{variationalequation}
\end{equation}  
and both $\mathcal{H}(\boldsymbol{u},\Theta)$ and $\mathcal{L}(\Theta)$ have the same global minimizer $\Theta^*$ if $\boldsymbol{u}\in\mathbb{U}$.

Problem \eqref{variationalequation} can be solved using an alternative minimization scheme  
\begin{equation}  
\begin{cases}
\boldsymbol{u}^{t+1}=\mathop{\mathrm{arg~min}}\limits_{\boldsymbol{u}(x)\in\mathbb{U}}\mathcal{H}(\boldsymbol{u},\Theta^{t}),\\
\Theta^{t+1}=\mathop{\mathrm{arg~min}}\limits_{\Theta}\mathcal{H}(\boldsymbol{u}^{t+1},\Theta).
\end{cases}
\label{altenative1}
\end{equation}  
It can be shown that this constraint optimization method is equivalent to the EM algorithm. 

After reinterpreting the EM algorithm with a constraint optimization framework, the GMM-EM model can be easily regularized by constraining the function $\boldsymbol{u}$, and get the following EMTV model:
\begin{equation}\label{EMTVModel}
\mathop{\mathrm{min}}\limits_{\boldsymbol{u}\in\mathbb{U},\Theta} -\int_\Omega\sum\limits_{i=1}^I\log[\alpha_ip_i(I(x);\boldsymbol{ \mu}_i,\boldsymbol{ \sigma}_i)]u_i(x)dx+\int_\Omega\sum\limits_{i=1}^Iu_i(x)\log u_i(x)dx+
\lambda\int_\Omega|\nabla\boldsymbol{u}|dx.
\end{equation}

\subsection{The global minimization of continuous multiphase partitioning method \cite{Bae2011}}
Since the TV regularization in \eqref{Potts} is not linear about $\bm u$, one can use the dual formulation of TV, that is 
$$\underset{\bm q_i\in \mathbb{Q}}{\max}\sum\limits_{i=1}^I\sum\limits_{j=1}^Ju_i(x_j)div(\bm q_i) (x_j)$$
$$\mathbb{Q}=\{\bm q_i\in C_0^1: ||\bm q_i(x_j)||_{2}\leqslant   1 \}.$$
This dual formula is linear with respect to $\bm u$ and the $\bm u$-subproblem has a close-formed solution.
However, the extra dual variable $\bm q$ is unknown and one has to iteratively solve it. One common choice for updating $\bm q$ is the projection gradient ascend method, and one can get the following iteration
\begin{equation}\label{td_tv}
\left\{
\begin{array}{rl}
u_i^{t+1}=&\Psi_i,\\
\bm  q_i^{t+1}=&\mathrm{Proj}_{\mathbb{Q}}\left(\bm  q_i^{t} -\tau_{\bm q} \nabla  u^{t+1}_i\right),\\
\end{array}
\right.
\end{equation}
where $\Psi_i$ is an indicative function defined by
\begin{equation*}
\Psi_i=\left\{
\begin{array}{rl}
1,&i\in\left\{\hat{i}:~\hat{i}=\arg\min\left\{c_1+div \bm q_1^{t},c_2+div \bm q_2^{t}\cdots,c_I+div \bm q_I^{t}\right\}\right\},\\
0,&\text{else}.
\end{array}
\right.
\end{equation*}
In the above derivation, we assume that the minimization of 
$$\arg\min\left\{c_1+div \bm q_1^{t},c_2+div \bm q_2^{t}\cdots,c_I+div \bm q_I^{t}\right\}$$ 
is unique. Otherwise, the $\Psi_i$ is not unique and this problem becomes very bad. However, one can numerically verify that such an alternating iteration fails to converge stably because of the binary of $\bm u_i^{t+1}$.
In \cite{Bae2011}, the authors proposed a smooth approximation version to get a stable algorithm.

Assume the iteration \eqref{td_tv} can reach a stationary point (it is unavailable in fact), then 
we put the solution $u^{*}_i$ obtained by \eqref{td_tv} into the energy \eqref{Potts}, with a dual representation of TV,
one can get the minimization of \eqref{Potts} is
\begin{equation}\label{std_TV}
\begin{array}{l}
\underset{\bm q\in \mathbb{Q}}{\max} \sum\limits_{j=1}^J \min\{c_1+div(\bm q_1),c_2+div(\bm q_2),\cdots,c_I+div(\bm q_I)\}\\
=\underset{\bm q\in \mathbb{Q}}{\max} \sum\limits_{j=1}^J -\max\{-c_1-div(\bm q_1),-c_2-div(\bm q_2),\cdots,-c_I-div(\bm q_I)\}.
\end{array}
\end{equation}

The $\max$ function appeared in the above energy is not smooth, it can be replaced by a smooth version $\max_{\varepsilon}$
defined by
 \begin{mydef}[$\max_{\varepsilon}$\cite{Bae2011}]
Given a vector $\mathbf{z}=(z_{1},z_{2},..z_{I})$, the $\max_{\varepsilon}$ operator is defined by
\begin{equation*}
\mathrm{max}_{\varepsilon}(\mathbf{z}):=\varepsilon\log\sum\limits_{i=1}^{I}\displaystyle\biggl. e^{\frac{z_{i}}{\varepsilon}}.
\end{equation*}
\end{mydef}
It is easy to check $\underset{\varepsilon\rightarrow 0}{\lim}~\mathrm{max}_{\varepsilon}(\textbf{z})=\max\{\bm z\}$ and $\max_{\varepsilon}$ is smooth. Thus the problem \eqref{std_TV} has a smooth version 
\begin{equation}\label{std_TV2}
\underset{\bm q\in \mathbb{Q}}{\max} \sum\limits_{j=1}^J -\mathrm{max}_{\varepsilon}\{-c_1-div(\bm q_1),-c_2-div(\bm q_2),\cdots,-c_I-div(\bm q_I)\}.
\end{equation}
The above equation is the intrinsic formulation used 
in \cite{Bae2011}. We can mathematically show this approximation in dual space is an entropic regularization in prime space. 
We need the following proposition to help us.
\begin{pro}\label{pro5}
Let
\begin{equation*}\label{f_p}
\mathcal{F}(\mathbf{z})=\mathrm{max}_{\varepsilon}(\textbf{z}),
\end{equation*}
then its Fenchel-Legendre transformation
\begin{equation*}\label{df_p}
\begin{array}{lll}
\mathcal{F}^{*}(\bm u)=&\max\limits_{\mathbf{z}}\{<\mathbf{z},\bm u>-\mathcal{F}(\mathbf{z})\}\\
&=\left\{
\begin{array}{lll}
\varepsilon\displaystyle\sum\limits_{i=1}^{I}u_{i}\log u_{i},& \bm u\in\mathbb{U},\\
+\infty,& else.
\end{array}
\right.
\end{array}
\end{equation*}
where $\mathbb{U}=\{\bm u=(u_{1},u_{2},...,u_{I}):0\leqslant u_i\leqslant 1,\sum\limits_{i=1}^{I}u_i=1\}$.
Moreover, $\mathcal{F}(\mathbf{z})$ is convex with respect to $\bm z$ and thus 
$$\mathcal{F}(\mathbf{z})=\mathcal{F}^{**}(\mathbf{z})=\max\limits_{\bm u\in\mathbb{U}}\left\{<\mathbf{z},\bm u>-\varepsilon\displaystyle\sum\limits_{i=1}^{I}u_i\log u_i\right\}.$$
\end{pro}
The proof of the proposition is a standard argument of convex optimization, we leave it to the readers.

Applying this proposition, $\max_{\varepsilon}$ can be formulated as an entropic regularization maximization problem. 

With simplifications, the problem \eqref{std_TV2}   becomes
\begin{equation}\label{std_TV22}
\underset{\bm q\in \mathbb{Q}}{\max} \sum\limits_{j=1}^J \underset{u\in\mathbb{U}}{\min}\left\{\sum_{i=1}^I (c_{i,j}+div(\bm q_i))u_{i,j}
+\varepsilon\sum\limits_{i=1}^{I}u_{ij}\log u_{ij}\right\}.
\end{equation}
Let us mention that the prime and dual relationship with entropic regularization was not given in \cite{Bae2011}, thus it is not easy to put the volume preserving in dual space if one followed the dual method in \cite{Bae2011}.

\subsection{The motivation of the proposed method}
Since the regularization in threshold dynamics method is not the exact length of region boundaries. One simple idea is 
to replace the regularizer with TV. However, this will significantly destroy the linearity property of the cost functional and the solution of $u_i$ is not a binary assignment problem, thus it cannot be solved by the discrete auction algorithm. Another choice is using the dual formulation of TV, thus this dual formula is linear with respect to $\bm u$ and the $\bm u$-subproblem can be still solved by auction algorithm. We can get the same iteration \eqref{td_tv} except that $\Psi_i$ is solved by the auction algorithm.

However, one can show that this iteration also fails to converge stably because of the binary of $ u^{t+1}_i$. What is more, the solution of threshold dynamics method is binary, it can not be directly applied to DCNN since the non-smoothness would lead to a gradient explosion in back propagation.

The goal of this paper is to propose a stable continuous smooth dual algorithm for TV regularized threshold dynamics volume constrained segmentation, and it can be directly enrolled into the DCNN layers. To obtain smoothness of $u_i$, according to the derivation in the above subsection, we would like to add another regularization term called entropy regularizer, which is closely connected with softmax.

%
%
\section{The proposed method}
In this section, we will first propose our volume constrained image segmentation model. Then we show that this proposed model can be interpreted as an entropic regularized optimal transport \cite{cuturi2016smoothed}, and it can be solved by
a stable dual method. Then, we will show it can be directly unrolled as a new network layers in Section 5.

\subsection{The proposed volume constrained image segmentation model}
Our model can be given as
\begin{equation} 
\begin{split}
\mathop{\mathrm{min}}\limits_{\boldsymbol{u}\in\mathbb{U}(\bm V) }& 
\sum\limits^I_{i=1}\sum\limits_{j=1}^Jc_{i}(x_j)u_i(x_j)+\varepsilon\sum\limits^I_{i=1}\sum\limits_{j=1}^Ju_i(x_j)\mathrm{\log}u_i(x_j)+\lambda\sum\limits^I_{i=1}\sum\limits^J_{j=1}e(x_j)||\nabla u_i(x_j)||. \\
\end{split}
\label{ProposedModel1}
\end{equation} 
Here the second term is a negative entropy term which can
enforce $\bm u$ to be smooth. $\varepsilon>0$ is a control parameter. $e(x_j)$ appeared in TV is an edge detection function such as $\frac{1}{1+\varepsilon^{'}|\nabla \bm (k*u)(x_j)|}.$

To be different from the Potts model \eqref{Potts} and threshold dynamics model \eqref{ThreshD}, the proposed model contains a negative entropy term, which is very important 
for the stability of the algorithm. Besides, similar as Potts model, the TV regularization is adopted in \eqref{ProposedModel1} 
, and it ensure that our method can exactly penalize the length of contour lines. What is more, $\boldsymbol{u}\in\mathbb{U}(\bm V)$ can guarantee our model to have a volume preserving property. Thus, the proposed model has the superiority of both Potts and threshold dynamic models.
More importantly, the new introduced negative entropy can produce a softmax operator, which can be directly used to construct a new layer in DCNN.  

For convenience, we rewrote \eqref{ProposedModel1} as a 
dual representation
\begin{equation} 
\begin{split}
& \mathop{\mathrm{min}}\limits_{\boldsymbol{u}\in\mathbb{U}(\bm V)}\mathop{\mathrm{max}}\limits_{\boldsymbol{q}\in\mathbb{Q}}\left\{
\sum\limits^I_{i=1}\sum\limits_{j=1}^J\left(c_{i}(x_j)+div(\bm q_i)(x_j)\right)u_i(x_j)+\varepsilon\sum\limits^I_{i=1}\sum\limits_{j=1}^Ju_i(x_j)\mathrm{\log}u_i(x_j)\right\},\\ 
\end{split}
\label{ProposedModel2}
\end{equation} 
where $$\mathbb{Q}=\left\{\bm q=(\bm q_1,\bm q_2,\cdots, \bm q_I),\bm q_i\in C_0^1: ||\bm q_i(x_j)||_2\leqslant   \lambda e(x_j) \right\}.$$

Since the introduction of the negative entropy, the dual problem of this problem becomes more difficult. However, the negative entropy is convex, it is easy to check \eqref{ProposedModel2} is convex problem if $\bm u >0$ (cost functional and constraint $\mathbb{U}(\bm V)$ are both convex), and
this problem would have an efficient solver.

Before to solve this problem, let us first give an optimal transport interpretation for our proposed model \eqref{ProposedModel2} from entropic regularized optimal transport.

\subsection{Optimal transport interpretation}
\subsubsection{Kantorovitch formulation and entropic regularized optimal transport problem}
Le us consider the discrete formulation of the Kantorovitch optimal mass transportation problem between a pair of  measures $\mathbf{a}\in \Sigma_{n,J}$ and $\mathbf{b}\in \Sigma_{m,J}$. Here the set $\Sigma_{n,J}:=\{\mathbf{ x}\in\mathbb{R}^n_+; \langle \mathbf{ x},\mathbf{1}_n\rangle=J\}$ is a simplex of histogram vectors (e.g. $\Sigma_{n,1}$ is a discrete probability simplex of $\mathbb{R}^n$). Thus the histograms are defined on $\mathbb{R}^n_+$ and $\mathbb{R}^m_+$ with the constraint that both masses are equal $\sum\limits_{i=1}^n a_i=\sum\limits_{j=1}^m b_j$. Considering a cost matrix $\mathbf{ C}\in\mathbb{R}^{n\times m}$, each element $\mathrm{\mathbf{C}}_{i,j}$ stands for the cost required to transfer a unit of mass from bin $i$ to bin $j$. The Kantorovitch formulation of optimal transport problem is
\begin{equation}    
\mathrm{\mathbf{L}}_{\mathrm{\mathbf{C}}}(\mathbf{a},\mathbf{b})=\mathop{\mathrm{min}}\limits_{\mathrm{\boldsymbol{u}}\in\mathrm{\mathbb{U}}(\mathbf{a},\mathbf{b})}\langle\mathrm{\boldsymbol{u}},\mathrm{\mathbf{C}}\rangle,
\label{KPproblem}
\end{equation}  
where 
$$\mathrm{\mathbb{U}}(\mathbf{a},\mathbf{b})=\{\mathrm{\boldsymbol{u}}\in\mathbb{R}^{n\times m}_+;\mathrm{\boldsymbol{u}}\mathbf{1}_m=\mathbf{ a},\mathrm{\boldsymbol{u}}^\top\mathbf{1}_n=\mathbf{ b}\}$$
is the set of couplings linking a pair of histograms $(\mathbf{ a},\mathbf{ b})$. The element $\mathrm{\boldsymbol{u}}_{i,j}$ represents the amount of mass transferred from bin $i$ to bin $j$, and the constraints account for the conservation of mass. The feasible set $\mathrm{\mathbb{U}}(\mathbf{a},\mathbf{b})$ is bounded and is a convex polytope. The Kantorovitch formulation aims at finding an optimal coupling minimizes the global transport cost and it is a linear program which usually not admits a unique optimal solution. 

Due to the optimal coupling of the Kantorovitch formulation is hard to compute in practical applications, especially for high dimension histograms such as image segmentation problem. An entropic regularized optimal transport \cite{Cuturi2013Sinkhorn, cuturi2014fast, cuturi2016smoothed} was proposed. By adding an entropic regularization to the original problem, one can get the entropic regularized Kantorovitch problem
\begin{equation} 
\mathrm{\mathbf{L}}_{\mathrm{\mathbf{C}}}^\varepsilon(\mathbf{a},\mathbf{b})=\mathop{\mathrm{min}}\limits_{\mathrm{\boldsymbol{u}}\in\mathrm{\mathbb{U}}(\mathbf{a},\mathbf{b})}\langle\mathrm{\boldsymbol{u}},\mathrm{\mathbf{C}}\rangle-\varepsilon\mathrm{\mathbf{ H}}(\mathrm{\boldsymbol{u}})
\label{EntropyProblem}
\end{equation} 
where the discrete entropy of a matrix $\boldsymbol{u}$ is defined as
$$\mathrm{\mathbf{ H}}(\mathrm{\boldsymbol{u}})=-\sum\limits_{i,j}\mathrm{\boldsymbol{u}}_{i,j}\log(\mathrm{\boldsymbol{u}}_{i,j}).$$
It is the classical optimal transport when $\varepsilon=0$. For $\varepsilon>0$, since the objective is a  $\varepsilon$-strongly convex function, the problem \eqref{EntropyProblem} has a unique optimal solution $\boldsymbol{u}_\varepsilon$.
Moreover, the solution $\boldsymbol{u}_\varepsilon$
has the following property:

\begin{pro}[(Convergence with $\varepsilon$ \cite{cuturi2016smoothed})]\label{pro1}
The unique solution $\boldsymbol{u}_\varepsilon$ of \eqref{EntropyProblem} converges to the optimal solution with maximal entropy within the set of all optimal solutions of the Kantorovich problem, namely
$$\mathrm{\boldsymbol{u}}_\varepsilon\mathop{\longrightarrow}\limits^{\varepsilon\rightarrow0}\mathop{\mathrm{argmin}}\limits_{\mathrm{\boldsymbol{u}}}\{-\mathrm{\mathbf{ H}}(\mathrm{\boldsymbol{u}});\mathrm{\boldsymbol{u}}\in\mathrm{\mathbb{U}}(\mathbf{a},\mathbf{b}),\langle\mathrm{\boldsymbol{u}},\mathrm{\mathbf{C}}\rangle=\mathrm{\mathbf{L}}_{\mathrm{\mathbf{C}}}^0(\mathbf{a},\mathbf{b})\},$$
so that in particular $\mathrm{\mathbf{L}}_{\mathrm{\mathbf{C}}}^\varepsilon(\mathbf{a},\mathbf{b})\mathop{\longrightarrow}\limits^{\varepsilon\rightarrow0}\mathrm{\mathbf{L}}_{\mathrm{\mathbf{C}}}^0(\mathbf{a},\mathbf{b})$. One also has
$$\mathrm{\boldsymbol{u}}_\varepsilon\mathop{\longrightarrow}\limits^{\varepsilon\rightarrow\infty}\frac{1}{J}\mathbf{a}\mathbf{b}^\top=\frac{1}{J}(a_ib_j)_{i,j},$$
where $J=\sum_ia_i=\sum_jb_j$.
\end{pro}
This proposition states that for a small regularization ($\varepsilon\rightarrow0$), the regularized solution converges to the original optimal transport coupling that has the highest entropy. In the opposite case ($\varepsilon\rightarrow\infty$), the regularized solution converges to the coupling with maximal entropy in the feasible set, i.e. the joint probability between the two prescribed marginals $\mathbf{a}, \mathbf{b}$. A refined analysis of this convergence with $\varepsilon$ is performed in \cite{cominetti1994asymptotic}. This regularized optimal transport problem \eqref{EntropyProblem} has an important advantage that its dual problem is a smooth convex optimization problem, thus can be solved using a simple alternating minimization scheme, which is very favorable for numerical implementation. 

\subsubsection{Interpreting the proposed model with  regularized optimal transport}

Now we try to interpret and solve our proposed model 
\eqref{ProposedModel2} with the framework of entropic 
regularized optimal transport theory. Firstly, let $\mathbf{a}
=(V_1,\cdots,V_I)^\top$, $\mathbf{b}=\mathbf{1}_J$ 
( according to the volume constrained image segmentation 
model, we have $\sum\limits_{i=1}^Ia_i=\sum\limits_{j=1}
^Jb_j=|\Omega|=J$, thus the two marginal histograms satisfy the 
constraint that the total masses are equal), and $\mathbf{C}
_{i,j}=c_i(x_j)$,  $\boldsymbol{u}_{i,j}=u_i(x_j)$, 
$i=1,\cdots,I$, $j=1,\cdots,J$.  Denote $\mathbf{D}_{i,j}
=div(\mathbf{q}_i)(x_j)$, then the problem 
\eqref{ProposedModel2} can be rewritten as
\begin{equation} 
\mathrm{\mathbf{L}}_{\mathrm{\mathbf{C}}}^{\varepsilon,
\lambda}(\mathbf{a},\mathbf{b})=\mathop{\mathrm{min}}
\limits_{\mathrm{\boldsymbol{u}}\in\mathrm{\mathbb{U}}
(\mathbf{a},\mathbf{b})}\langle\mathrm{\boldsymbol{u}},
\mathrm{\mathbf{C}}+\bm 
D\rangle-\varepsilon\mathrm{\mathbf{ H}}
(\mathrm{\boldsymbol{u}})
\label{OurModel}
\end{equation} 
if the dual variable $\bm q_i$ is given.
Therefore the $\bm u$-subproblem of the proposed model is an 
entropic regularized optimal transport problem when we 
apply alternating optimization scheme. In the $\bm u$-
subproblem, we consider the volume constraint as a measure
$\mathbf{a}$, and image segmentation condition 
as another measure $\mathbf{b}$. Thus the volume constrained 
image segmentation problem can be seen as a 
transportation  problem between the volume constraints and 
image segmentation condition. The element $
\mathrm{\boldsymbol{u}}_{i,j}$ represents the probability 
that pixel $x_j$ should be assigned to phase $i$. The sum 
constraints on rows $\mathrm{\boldsymbol{u}}\mathbf{1}
_J=\mathbf{ a}$ represent the volume constraints in the 
image segmentation problem, and the sum constraints on 
columns $\mathrm{\boldsymbol{u}}^\top\mathbf{1}
_I=\mathbf{ b}$ represents that each pixel is allocated to 
phases with probability sum of 1. Therefore, under the cost 
matrix $\mathrm{\mathbf{C}}+\mathrm{\mathbf{D}}$ which 
contains the information of similarity and TV regularization, 
the image segmentation model aims to find an optimal 
coupling $\mathrm{\boldsymbol{u}}$ matching the pixels 
and volumes that minimizes the global transport cost. Figure 
1 gives an intuitive explanation of this transport problem, 
where we set $\mathbf{a}=[2,5,3]^\top$, $\mathbf{b}
=\mathbf{1}_{10}$, $\mathrm{\mathbf{C}}_{i,j}=(2i-j)^2$,  $
\mathrm{\mathbf{D}}=\mathbf{0}$. The red and blue points 
represent the measure $\mathbf{a}$ and $\mathbf{b}$ 
respectively, and black points according to the optimal 
coupling $\mathrm{\boldsymbol{u}}$, and their size is 
proportional to their value. Figure 1 also displays the 
influence of $\varepsilon$ on $\mathrm{\boldsymbol{u}}$, as 
$\varepsilon$ increases, the optimal coupling becoming less 
and less sparse, which will be discussed in Section 6 
detailedly.

 \begin{figure}
\centering
\includegraphics[width=0.95\linewidth]{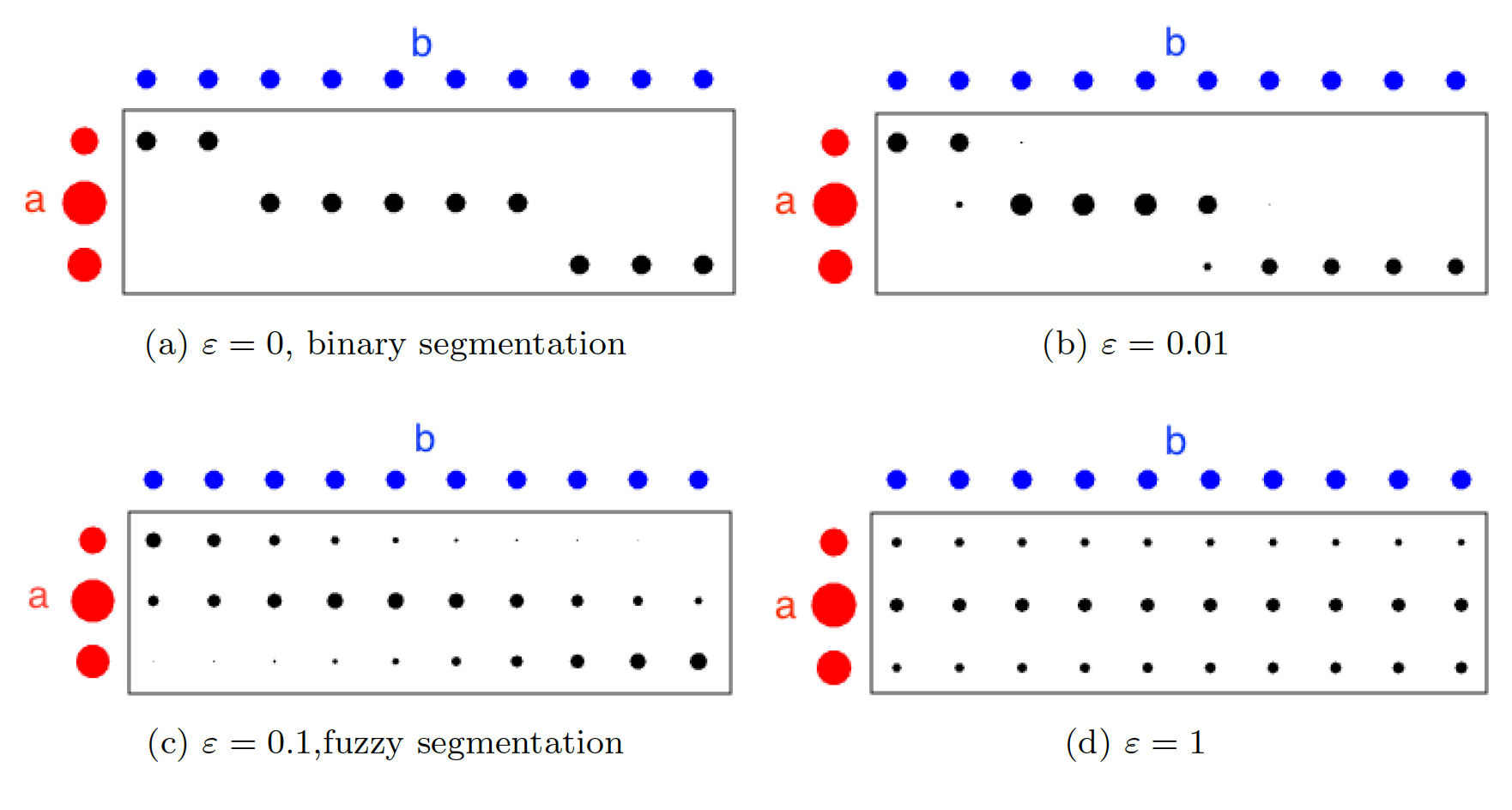}
%
\caption{Exhibition of optimal transportation for volume constrained image segmentation, and impact of $\varepsilon$ on the optimal couplings. The size of points is proportional to their values.}
\end{figure}

\subsection{The dual algorithm}
As the former analysis, our model can be solved by continuous dual algorithm.
The  problem \eqref{ProposedModel2} can be solved by applying the following alternating method:
\begin{eqnarray}  
\boldsymbol{u}^{t+1}&=&\mathop{\mathrm{arg~min}}\limits_{\mathrm{\boldsymbol{u}}\in\mathbb{U}(\bm V)}\langle\mathrm{\boldsymbol{u}},\mathrm{\mathbf{C}}+div(\bm q^{t})\rangle-\varepsilon\mathrm{\mathbf{H}}(\mathrm{\boldsymbol{u}}),
\label{PIter}\\
\mathbf{q}^{t+1}&=&\mathop{\mathrm{arg~max}}\limits_{\bm q \in\mathbb{Q}}\langle\boldsymbol{u}^{t+1}, div(\bm q)\rangle.
\label{DIter}
\end{eqnarray}

Let's first consider the $\bm q$-subproblem \eqref{DIter}. This problem can be easily solved by the Chambolle projection gradient method \cite{Chambolle2004}, one can obtain 
\begin{equation}  
\mathbf{q}_i^{t+1}(x_j)=\mathrm{Proj}_\mathbb{Q}\left(\mathbf{q}_i^{t}(x_j)-\tau_{\bm q}\nabla u_i^{t+1}(x_j)\right),
\label{DNew}
\end{equation}  
where 
$$\mathrm{Proj}_\mathbb{Q} \left(\bm q_i(x_j)\right)=\frac{\lambda e(x_j)\bm q_i(x_j)}{\mathrm{max}\{\|\bm q_i(x_j)\|_2,\lambda e(x_j)\}},$$
and $\tau_{\bm q}$ is a small time step.

In the next, let us consider the subproblem \eqref{PIter}. 
To simplify notations,  we let $\mathbf{D}_{i,j}=div(\mathbf{q}_i^{t})(x_j), \bm C_{i,j}=c_i(x_j),u_i(x_j)=u_{i,j}$. 
Then we have $\bm u$-subproblem
\begin{equation}  
\mathrm{\mathbf{L}}_{\mathrm{\mathbf{C}},\mathbf{D}}^{\varepsilon}(\mathbf{V})=\mathop{\mathrm{min}}\limits_{\mathrm{\boldsymbol{u}}\in\mathbb{U}(\bm V)}\langle\mathrm{\boldsymbol{u}},\mathrm{\mathbf{C}}+\bm D\rangle-\varepsilon\mathrm{\mathbf{H}}(\mathrm{\boldsymbol{u}}).
\label{PProblem}
\end{equation}  
Follow the work of \cite{cuturi2016smoothed}, we can deduce an equivalent dual formulation to problem \eqref{PProblem}:

\begin{pro}\label{pro2}
For $\varepsilon\geqslant0$, one has the equivalent dual formulation
\begin{equation} 
\mathrm{\mathbf{L}}_{\mathrm{\mathbf{C}},\mathbf{D}}^{\varepsilon}(\bm V)=\mathop{\mathrm{max}}\limits_{\bm f\in\mathbb{R}^I,\bm g\in\mathbb{R}^J}\langle\bm f,\bm V\rangle+\langle\bm g,\bm 1\rangle+B_{\varepsilon}(\mathrm{\mathbf{C}}+\mathbf{D}-\bm f\oplus\bm g),
\label{DualProblem1}
\end{equation}
where $\mathbf{S}\in\mathbb{R}^{I\times J}$
$$B_{0}(\mathbf{S})=-\iota_{\mathbb{R}^{I\times J}_+}(\mathrm{\mathbf{S}})\quad\mathrm{and}\quad\forall\varepsilon>0,~ B_{\varepsilon}(\mathrm{\mathbf{S}})=-\varepsilon\sum\limits_{i,j}e^{-\mathrm{\mathbf{S}}_{i,j}/\varepsilon-1},$$
and $$(\bm f\oplus\bm g)_{i,j}= f_i+ g_j.$$
Moreover, the prime variable $\bm u$ and dual variables $\bm f,\bm g$ are connected by
\begin{equation}  
u_{i,j}=e^{\frac{-\mathbf{C}_{i,j}-\mathbf{D}_{i,j}+f_i^{*}+g_j^{*}}{\varepsilon}-1}, 
\label{POptimal}
\end{equation}
where $\bm f^{*},\bm g^{*}$ are the related maximizers.
\end{pro}
The proof of this proposition can be found in Appendix \ref{proofpro2}.

Inspired by the $c$-transform theory of optimal transport, we give a variant definition of $c$-transforms to our problem. For $\varepsilon\geqslant0$, define
\begin{equation} 
\forall~\bm f\in\mathbb{R}^I,\forall j\in[1,2,\cdots,J],~f_j^{c,\varepsilon}=\mathrm{min}_{\varepsilon}(\mathrm{\mathbf{C}}_{\cdot,j}+\mathbf{D}_{\cdot,j}-\bm f),
\label{CTransform1}
\end{equation} 
\begin{equation} 
\forall~\bm g\in\mathbb{R}^J,\forall i\in[1,2,\cdots,I],~g_i^{\bar{c},\varepsilon}=\varepsilon\log(V_i)+\mathrm{min}_{\varepsilon}(\mathrm{\mathbf{C}}_{i,\cdot}+\mathbf{D}_{i,\cdot}-\bm g),
\label{CTransform2}
\end{equation}
where
$$\mathrm{min}_0(\mathbf{z}):=\mathrm{min}~\mathbf{z}=\underset{i}{\mathrm{min}}\{z_1,z_2,\cdots,z_I\}\quad\mathrm{and}\quad\mathrm{min}_\varepsilon\mathbf{z}:=-\varepsilon\log\sum\limits_ie^{-z_i/\varepsilon-1}.$$
One can check that for a fixed $\bm f$, 
$\bm f^{c,\varepsilon}$ is the maximizer of \eqref{DualProblem1} with respect to $\bm g$.
With the similar way, for a fixed $\bm g$, 
$\bm f=\bm g^{\bar{c},\varepsilon}$ is the maximizer of \eqref{DualProblem1} with respect to $\bm f$.
This fact produces the Sinkhorn iteration  \cite{Cuturi2013Sinkhorn} to solve dual variable, 
$$\bm g^{l}=(\bm f^{l})^{c,\varepsilon}~~\mathrm{and}~~~\bm f^{l+1}=(\bm g^{l})^{\bar{c},\varepsilon}.$$
Let us emphasize that 
this alternating $c$-transforms iterative scheme would not converge when $\varepsilon=0$ because the dual problem \eqref{DualProblem1} is not smooth. Indeed, we have the following inequalities
$$\langle\bm f,\bm V\rangle+\langle\bm f^{c,\varepsilon},\bm 1\rangle\leqslant\langle\bm f^{c,\varepsilon;\bar{c},\varepsilon},\bm V\rangle+\langle\bm f^{c,\varepsilon},\bm 1\rangle\leqslant\langle\bm f^{c,\varepsilon;\bar{c},\varepsilon},\bm V\rangle+\langle\bm f^{c,\varepsilon;\bar{c},\varepsilon;c,\varepsilon},\bm 1\rangle\leqslant\cdots$$
 which would reach a stationary point $(\bm f^{l+1},\bm g^{l+1})=(\bm f^{l},\bm g^{l})$ for $l=2$ since $\bm f^{c,\varepsilon;\bar{c},\varepsilon;c,\varepsilon}=\bm f^{c,\varepsilon}$. Therefore, the corresponding Sinkhorn algorithm fails, and the problem usually has no unique solution in this case. While for $\varepsilon>0$, the objective is a $\varepsilon$-strongly convex function, problem \eqref{PProblem} has a unique optimal solution, and the unique solution converges to the optimal solution with maximal entropy of the problem with $\varepsilon=0$. In addition, the corresponding Sinkhorn algorithm also converges linearly \cite{Franklin1989On}. 
   
 Once we use formula \eqref{DualProblem1} to get a convergence solution $(\bm f^*, \bm g^*)$, then the $\bm u$-subproblem can be recovered by \eqref{POptimal} .
 
The above analysis show that we can solve $\bm u$-subproblem by updating two dual variables $\bm f,\bm g$. In the next, we show these two dual variables are dependent and thus we only to solve one dual variable problem. Indeed, substituting \eqref{CTransform1} into \eqref{CTransform2}, one can obtain the iteration
\begin{equation}
f_i^{t+1}=\varepsilon\log V_i-\varepsilon\log\sum\limits_{j=1}^J\frac{e^{\frac{-\bm C_{i,j}-\bm D_{i,j}}{\varepsilon}}}{\sum\limits^I_{\hat{i}=1}e^{\frac{-\bm C_{\hat{i},j}-\bm D_{\hat{i},j}+f_{\hat{i}}^{t}}{\varepsilon}}}.
\label{}
\end{equation}
In numerical computation, we will use a stabilization trick to avoid overflow for small values of $\varepsilon$. This trick suggests to add the previously computed dual variable, which leads to the following stabilized iteration
\begin{equation}
f_i^{t+1}=\varepsilon\log V_i-\varepsilon\log\sum\limits_{j=1}^J\frac{e^{\frac{-\bm C_{i,j}-\bm D_{i,j}+f_i^{t}}{\varepsilon}}}{\sum\limits^I_{\hat{i}=1}e^{\frac{-\bm C_{\hat{i},j}-\bm D_{\hat{i},j}+f_{\hat{i}}^{t}}{\varepsilon}}}+f_i^{t}.
\label{}
\end{equation}

%

This is a simple iteration scheme for dual variable $\bm f$. In addition, by using formulation \eqref{POptimal} and the condition $$\sum\limits^I_{i=1}u_{i,j}=e^\frac{g_j}{\varepsilon}\sum\limits^I_{i=1}e^{\frac{-\bm C_{i,j}-\bm D_{i,j}+f_i}{\varepsilon}-1}=1,$$ we can obtain
\begin{equation}  
u_{i,j}=\frac{e^{\frac{-\bm C_{i,j}-\bm D_{i,j}+f_i}{\varepsilon}}}{\sum\limits^I_{\hat{i}=1}e^{\frac{-\bm C_{\hat{i},j}-\bm D_{\hat{i},j}+f_{\hat{i}}}{\varepsilon}}}.
\label{uvalue}
\end{equation}  
Thus we can recover the prime variable $\bm u$ when we get a converged $\bm f$.
And the usual way to segment an image is to classify pixel at $x_j$ into the phase with the largest probability:
$$label(h(x_j))=\mathop{\mathrm{arg~max}}\limits_{1\leqslant i\leqslant I}\{\boldsymbol u_{i}(x_j)\}.$$
 
 Combine the three $\bm u,\bm q,\bm f$ subproblem solvers, 
 we can get an algorithm which is summarized in algorithm 
 \ref{alg1}.
 
 \begin{algorithm}
\caption{Proposed volume constrained image segmentation.}\label{alg1}
\KwIn{Image $h$, the total number of phases, volume constraint vector $V=(V_1,\cdots,V_I)^\top$, parameters $\varepsilon$, $\lambda$, $\tau_{\bm q}$.}
\KwOut{Segmentation function $\boldsymbol{u}$.}
\textbf{Initialization:} $\mathbf{q}^0=\mathbf{0},\bm f^0=0$\;
\For{$t=0,1,2,\cdots,T$}{
1. Compute the similarity $c_i(x_j)$.\\
2. Softmax image segmentation:
$$u^{t+1}_i(x_j)=\frac{e^{\frac{-c_{i}(x_j)-div\mathbf{q}_i^{t}(x_j)+f_i^{t}}{\varepsilon}}}{\sum\limits^I_{\hat{i}=1}e^{\frac{-c_{\hat{i}}(x_j)-div\mathbf{q}_{\hat{i}}^{t}(x_j)+f_{\hat{i}}^{t}}{\varepsilon}}}.$$\\
3. Regularization step: \\
\begin{equation*}  
\mathbf{q}_i^{t+1}(x_j)=\mathrm{Proj}_\mathbb{Q}\left(\mathbf{q}_i^{t}(x_j)-\tau_{\bm q}\nabla u_i^{t+1}(x_j)\right).
\label{DNew}
\end{equation*} 
4. Volume perserving step: \\ 
\begin{equation*}  
f_i^{t+1}=\varepsilon\log V_i-\varepsilon\log\sum\limits_{j=1}^J\frac{e^{\frac{-c_{i}(x_j)-div\mathbf{q}_i^{t+1}(x_j)+f_i^{t}}{\varepsilon}}}{\sum\limits^I_{\hat{i}=1}e^{\frac{-c_{\hat{i}}(x_j)-div\mathbf{q}_{\hat{i}}^{t+1}(x_j)+f_{\hat{i}}^{t}}{\varepsilon}}}+f_i^{t}.
\end{equation*}  
5. Convergence check. If it is not converged, go back to step 1 to update $c_i(x_j)$ if one would like to update the similarity, or go to step 2. Else, go to step 6.\\
6. Label function:
$$label(h(x_j))=\mathop{\mathrm{arg~max}}\limits_{1\leqslant i\leqslant I}\{\boldsymbol u_{i}(x_j)\}.$$
}
\Return Segmentation result: label function.
\end{algorithm}

\section{Relationships with other methods}
In this section, we will clarify the relationships among our proposed model and other methods aforementioned in Section 2. This is summarized in \tablename\ref{tab:relations}.

\begin{table}[htp]
\caption{Relationships among the proposed method and other image segmentation methods('Y', 'N' means 'Yes', 'No', respectively).}\label{tab:relations}
\begin{center}
\begin{tabular}{p{3.2cm}<{\centering}|p{1.5cm}<{\centering}|p{2cm}<{\centering}|p{1.5cm}<{\centering}|p{1.5cm}<{\centering}}
\hline
 Method & Model formulation & Boundaries smothness& Entropy regularizer & Volume preserving\\
\hline
\hline
Potts\cite{Potts1952} & \eqref{Potts} & Y & N & N\\
\hline
Auction dynamics\cite{jacobs2018auction} & \eqref{ThreshD} & N(approximate) & N & Y\\
\hline
EMTV\cite{liu2012expectation} & \eqref{EMTVModel} & Y & Y($\varepsilon=1$) & N\\
\hline
Global minimization of continuous multiphase partitioning method\cite{Bae2011} & \eqref{std_TV22} & Y & Y & N\\
\hline
Proposed & \eqref{ProposedModel1}or\eqref{ProposedModel2} & Y & Y & Y\\          
\hline
\end{tabular}
\end{center}
\label{}
\end{table}

Compared with Potts model\cite{Potts1952}, our proposed model integrates entropy regularizer, which is very important for the stability of the algorithm. In addition, we added the volume constraints to increase the segmentation accuracy.

Compared with the auction dynamics scheme (the threshold dynamics volume constrained segmentation)\cite{jacobs2018auction}, it is not difficult to notice that our proposed model integrates TV and entropic regularization, which can accurately penalize the length of the boundaries and is stable in numerical implementation. 

Auction dynamics scheme \cite{jacobs2018auction} extends the threshold dynamics to multiphase volume constrained curvature motion. This method is based on a variational framework for the MBO algorithm developed in \cite{esedoglu2015threshold}. They showed that the minimization problem is equivalent to an assignment task which is a combinatorial optimization problem. Then they choose a variant of the auction algorithm developed in \cite{bertsekas1979distributed} to solve the related assignment problem. Thus the auction dynamics scheme consists of alternating two steps: convolution with a kernel, and assigning set memberships via auction. 

From the previous discussion, our model \eqref{ProposedModel2} with $\varepsilon=0$ does not converge in alternating $\boldsymbol{u},\bm q$ and $\bm f$. In this case, one can solve it by using the auction dynamics scheme proposed in \cite{jacobs2018auction}. However, the TV regularizer cannot be adopted since it is not smooth and the algorithm fails to converge, which also shows the importance of entropy regularizer. And for small $\varepsilon$, the result of our proposed algorithm converges to auction dynamics's. 

We also show that our model is essentially related to the EMTV \cite{liu2012expectation}. From \eqref{EMTVModel}, it is easy to find that the EMTV model is a special case of the proposed model when the entropic regularization parameter $\varepsilon=1$ and without volume constraint. In this sense, our proposed model is a volume preserving EM types
image segmentation. And also integrates the advantages of both variational and statistics methods.

The global minimization of continuous multiphase partitioning method \cite{Bae2011} is also closely related to our method. It is not difficult to notice that formulation \eqref{std_TV22} is the proposed model \eqref{ProposedModel2} without volume constraint. Thus, the proposed model is an extension of the continuous multiphase partitioning method \cite{Bae2011} on volume preserving and deep learning (in the next section).

\section{Volume Preserving and TV regularized softmax (VPTV-softmax) layer for DCNN}
\subsection{Proposed VPTV-softmax}
The DCNN for image segmentation can be mathematically formulated as 
$$\bm u=\mathcal{N}_{\bm\Theta}(h),$$
where $\mathcal{N}_{\bm\Theta}$ is a DCNN parameterized by $\bm\Theta$.
Let $h=\bm c^0$, then the operator $\mathcal{N}$ has a special structure
which has the following recursive relationship:
\begin{equation}\label{eq:nn1}
\left\{
\begin{array}{rl}
\bm o^k=&\mathcal{T}_{  \Theta^{k-1}}(\bm c^{k-1}),\\
\bm c^{k}=&\mathcal{A}^{k}(\bm  o^k),\\ 
\end{array}
~~k=1,2,\cdots,K.
\right.
\end{equation}
Here $\mathcal{T}_{  \Theta^{k-1}}( \bm c)= \mathcal{W}^{k-1} \bm c + \bm  b^{k-1}$ is an affine transformation parameterized
by $ \Theta^{k-1}=\{\mathcal{W}^{k-1}, \bm  b^{k-1}\}$, and 
$\mathcal{A}^{k}$ is an activation function (e.g. sigmoid, softmax, ReLU \emph{etc.}) or sampling (e.g. downsampling, upsampling \emph{etc.}), $K$ stands for layers which is related the depth of the networks. The output of the network are $\mathcal{N}_{\bm\Theta}( h)=\bm c^{K}$, and the parameter set $\bm\Theta=\bigcup_{k=1}^K\{\Theta^{k-1}\}.$

Usually, the softmax activation function in DCNN for image segmentation is
 $$[\mathcal{A}^K(\bm o^K)]_{i,j}=[\text{softmax}(\bm o^K)]_{i,j}=\frac{\exp(o^K_{i,j})}{\sum_{\hat{i}=1}^I\exp(o^K_{\hat{i},j})}.$$
where $\bm o^K$ is a feature extracted by DCNN.
Compared with the above softmax formulation, the formulation $u_i^{t+1}(x_j)$ in our Algorithm \ref{alg1}
is just a modified softmax, namely
\begin{equation*}
u_i^{t+1}(x_j)=\text{softmax}\left(\frac{-c_{i}(x_j)-\mathrm{div}\mathbf{q}_i^{t}(x_j)+f_i^{t}}{\varepsilon}\right).
\end{equation*}
But unlike the classical softmax, our proposed VPTV-softmax contains dual variables $\bm q$ and $\bm f$ for spatial regularization and volume preserving. 

\begin{figure}
\includegraphics[width=0.95\linewidth]{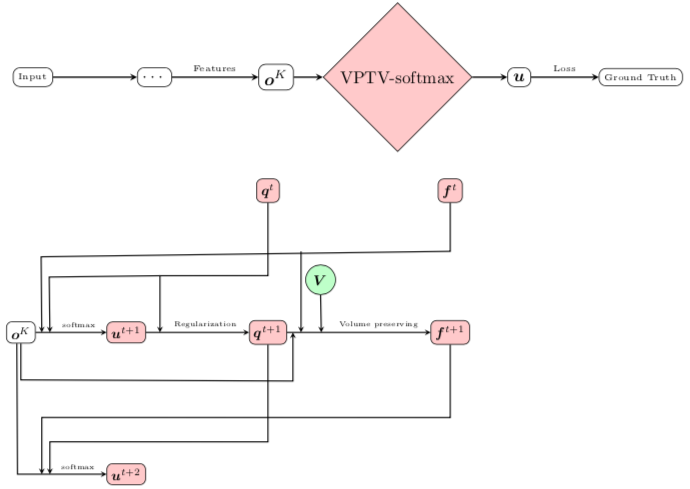}
\caption{The network architecture of the proposed VPTV-softmax block. The first picture is a flowchart of the VPTV-softmax in the DCNN, and the second one is the details structures of VPTV-softmax unrolled by one iteration in Algorithm \ref{alg1}.
}\label{fig:net}
\end{figure}

Thus, inspired by the proposed variational method and our previous work \cite{Jia2019}, we can modified the last layer of DCNN \eqref{eq:nn1} as 
\begin{equation}\label{eq:rnn}
\left\{
\begin{array}{ll}
\bm o^K=&\mathcal{T}_{ \Theta^{K-1}}( \bm c^{K-1}),\\
\bm c^{K}=&\mathcal{A}^K( \bm o^K)=\underset{\bm u\in \mathbb{U}(\bm V, \bm 1)}{\arg\min}\left\{-<\bm u, \bm o^K> + \varepsilon<\bm u,\log \bm u> +\lambda \text{TV}(\bm u) \right\}, \\
\end{array}
\right.
\end{equation}
The second optimization problem in the \eqref{eq:rnn} is just the proposed model \eqref{ProposedModel1} by replacing $\bm c$ with $-\bm o^K$,
and we have known that it has a smooth close-formed solution which is  related to softmax,
thus the back propagation for neural network is feasible in our problem. 
Therefore, the proposed algorithm can be unrolled as a new VPTV-softmax layer
as shown in \figurename \ref{fig:net}
\subsection{Implementation details}
The each step of the forward-propagating in the proposed VPTV-softmax has a close-formed formulation and it is smooth, thus the back-propagating is feasible. However, it is well-known that this gradient based algorithm is converged with a linear convergence rate, and it needs many inner iterations for dual variable $\bm q$ and $\bm f$, thus the related back-propagating would be time-consuming. Besides, such many layers would cost a lot of 
computational resource, especially for back propagation.
For computational efficiency, in this paper, we adopt a quasi dropout implementation. This implies that we drop some less important layers to avoid back-propagating in training but keep them in the forward-propagating of test.

\begin{figure}
\includegraphics[width=1.0\linewidth]{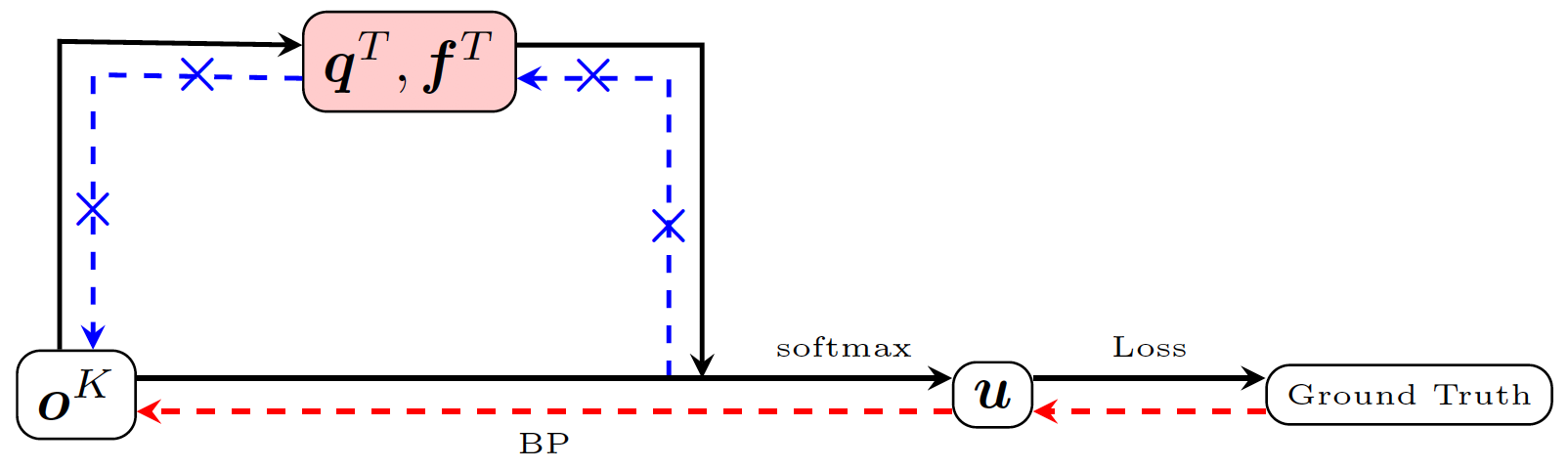}
\caption{The quasi dropout implementation of the proposed VPTV-softmax. Real and dash lines stand for forward and back propagating, respectively.}\label{fig:dropout}
\end{figure}

Let us analyze the algorithm \ref{alg1}, 
we can find that the final output $\bm u$ is just a softmax output by integrating the dual variable $\bm q$ (for spatial regularization) and $\bm f$ (for volume preserving). When we get the final $\bm q^T$ and $\bm f^T$, where $T$ is the last iteration number in algorithm \ref{alg1}, 
the output of the VPTV-softmax is just a softmax by combing features $\bm C, \bm q^T$ and $\bm f^T$. 
Thus there are two paths of back-propagating for layers $\bm C$, one is the classic softmax with inputs $\bm C, \bm q^T$ and $\bm f^T$, and the other is connections of $\bm q^T$ and $\bm f^T$. In real implementation, we drop the second back-propagating path
and empirically find that it does not effect the final results too much. But it can greatly improve the computational efficiency.
The implement details of the algorithm can be found in \figurename \ref{fig:dropout}. In this figure, the real lines stand for forward-propagating, and the dash lines represent back-propagating. The blue dash line (error propagation path) was set to zero when the network is  training.
\section{Numerical experiments}
In this section, we will illustrate the performance of our proposed model and its segmentation results through some numerical experiments. In the experiments, the dual variable $\bm q^0, \bm f^0$ are both initialized to $0$. The TV parameter $\lambda$ and entropy regularization parameter $\varepsilon$ are image-dependent. The choice criterion is that the larger $\lambda$ and smaller $\varepsilon$  for smoother boundaries and  binary segmentation, respectively. All the intensity of the images are normalized in $[0,1]$. Without specifically stated, the step size set as $\tau_{\bm q}=0.5*\varepsilon$, 
and the convergence standard is $\|\boldsymbol{u}^{t+1}-\boldsymbol{u}^{t}\|<10^{-3}$. 

\begin{figure}
\centering
\includegraphics[width=1.0\linewidth]{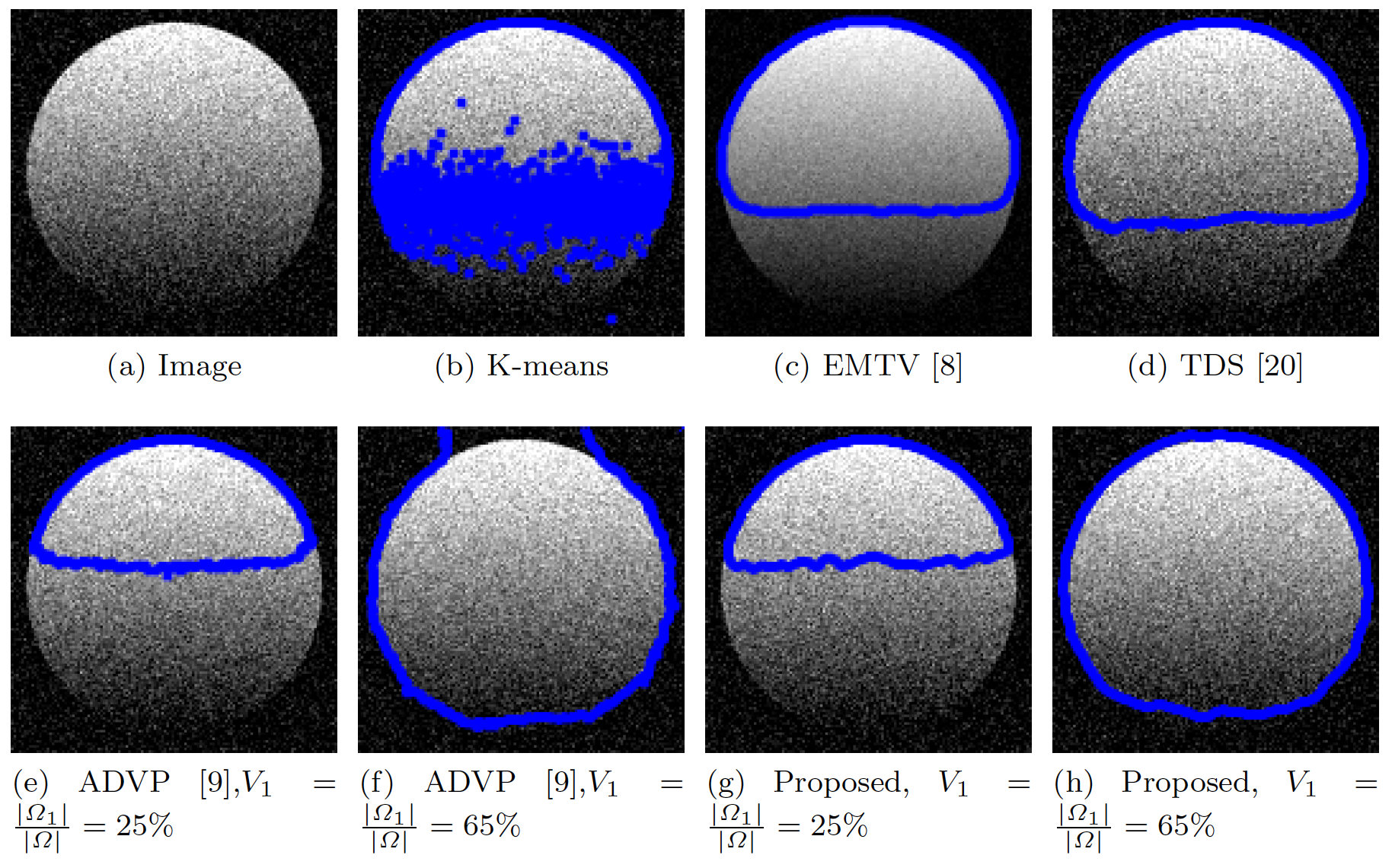}
%
\caption{Segmentation results of noisy and inhomogenous intensity.}\label{fig:exp1}
\end{figure}

The first experiment is a toy example is to compare the segmentation results by k-means, EMTV \cite{liu2012expectation}, threshold dynamics segmentation (TDS) \cite{Liu2011}, 
Auction Dynamics Volume Preserving (ADVP) \cite{jacobs2018auction} and our proposed model.
The test image is a synthetic image consisting of a solid black background and a circle with inhomogenous intensity. 
To test the robustness of the algorithm, the Gaussian white noise with variance 0.01 is added to the image, as shown in
\figurename\ref{fig:exp1}. The results with different algorithm are displayed in this figure.
In this experiment, the similarity $c_i({x_j})$ is to set as $(h(x_j)-m_i)^2$ where $m_i$ is the mean of the intensity given by the initial k-means.
Compared to k-means, EMTV and TDS provide smooth region boundaries due to the regularization. But they fail to give the circle contour 
because of inhomogenous intensity. The volume constraint can ensure that the algorithms can obtain some  special requirements for segmentation. As can be seen from this figure, when the volume constraint is $ V_1=\frac{|\Omega_1|}{|\Omega|}=25\%$, both of ADVP and the proposed algorithm just can segment some parts of the circle. It is reasonable due to a bad volume constraint.
When the ratio $ V_1$ is increasing to $65\%$, both of them can produce better results. However, the proposed method have slightly high segmentation accuracy because of existing of TV. This simple numerical example shows the superiority of the proposed algorithm.

\begin{figure}
\centering
\includegraphics[width=1.0\linewidth]{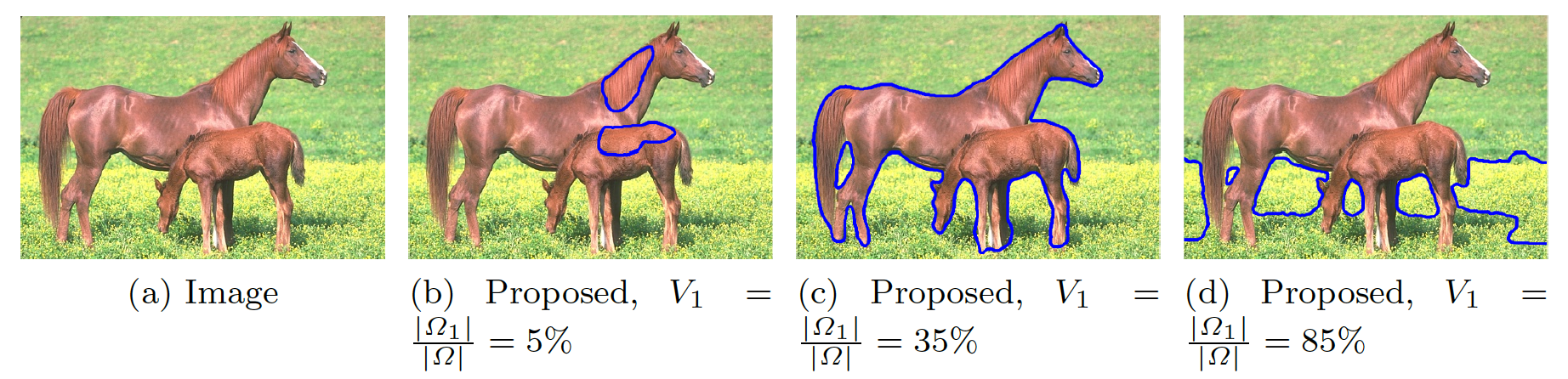}
\caption{Segmentation results of a natural image.}\label{fig:exp2}
\end{figure}

\begin{figure}
\centering
\includegraphics[width=0.9\linewidth]{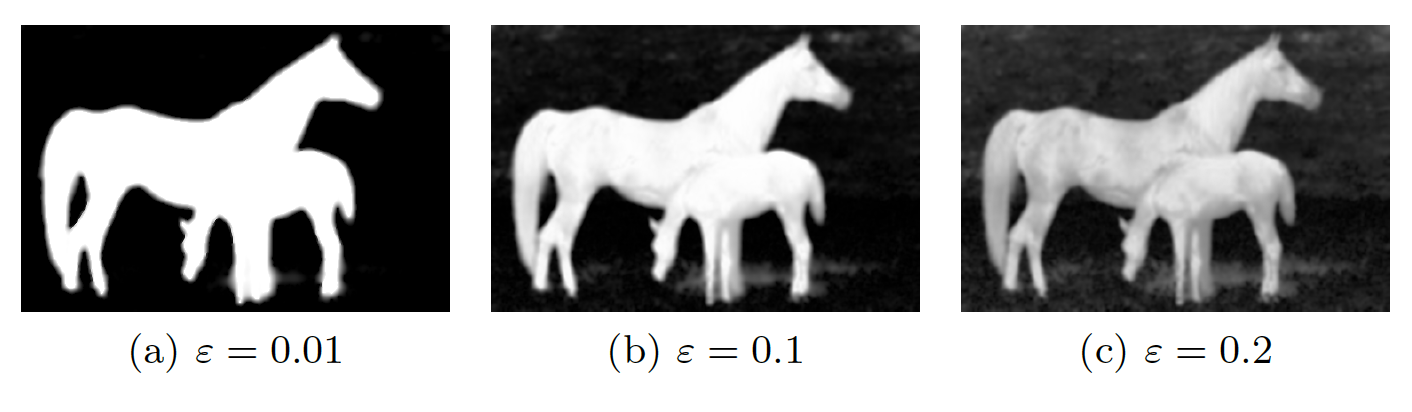}
\caption{The $\bm u$ affected by the entropic regularization parameter $\varepsilon$.}\label{fig:exp2:eps}
\end{figure}

The second numerical experiment is to test the performance of the algorithm on natural images.
For color image, the similarity can be 
$$c_{i}(x_j)=(\bm h(x_j)-\bm m_i)^\top\bm\Sigma_i^{-1}(\bm h(x_j)-\bm m_i), i=1,2,\cdots,I; j=1,2,\cdots,J,$$
where $\bm m_i, \bm\Sigma_i$ are the mean and covariance matrix of $i$-th class.
As shown in \figurename\ref{fig:exp2}, we classify this horse image into 2 classes with different volume constraints.
In the first case, the volume constraint is chosen as $V_1=\frac{|\Omega_1|}{|\Omega|}=5\%$, we find that the proposed 
algorithm can only separate the horsehairs from the image. In the second case, we change the volume ratio as $V_1=\frac{|\Omega_1|}{|\Omega|}=35\%$, this algorithm can automatically extract the horses well. When the ratio increases to $V_1=\frac{|\Omega_1|}{|\Omega|}=85\%$,
the dense yellow flowers can be detected by the proposed method. This experiment implies that the volume preserving term can 
segment different objects from an image if we know the volume prior. Let us emphasize that the entropic regularization parameter $\varepsilon$
would affect the volume preserving. The segmentation function $\bm u$ would be more and more smooth with the increasing of $\varepsilon$.
In these cases, the volume constraint is just an approximation. We should try to take small $\varepsilon$ if one would like to a strict volume constraint.
As it can be seen from \figurename \ref{fig:exp2:eps}, the $\bm u$ is nearly binary when $\varepsilon=0.01$, but it nearly a gray intensity image
when $\varepsilon=0.2$. This is the results under volume constraint $V_1=\frac{|\Omega_1|}{|\Omega|}=35\%.$
For computational stability, the $\varepsilon$ should be large. But for binary segmentation, the $\varepsilon$ should be small
as possible. As mentioned earlier, too small $\varepsilon$ would lead to the algorithm fails to converge. For image segmentation, it seems that
we chose $\varepsilon=0.01$ is a good balance. In this paper, without specific statements, we set $\varepsilon=0.01$. 

\begin{figure}
\centering
\includegraphics[width=1.0\linewidth]{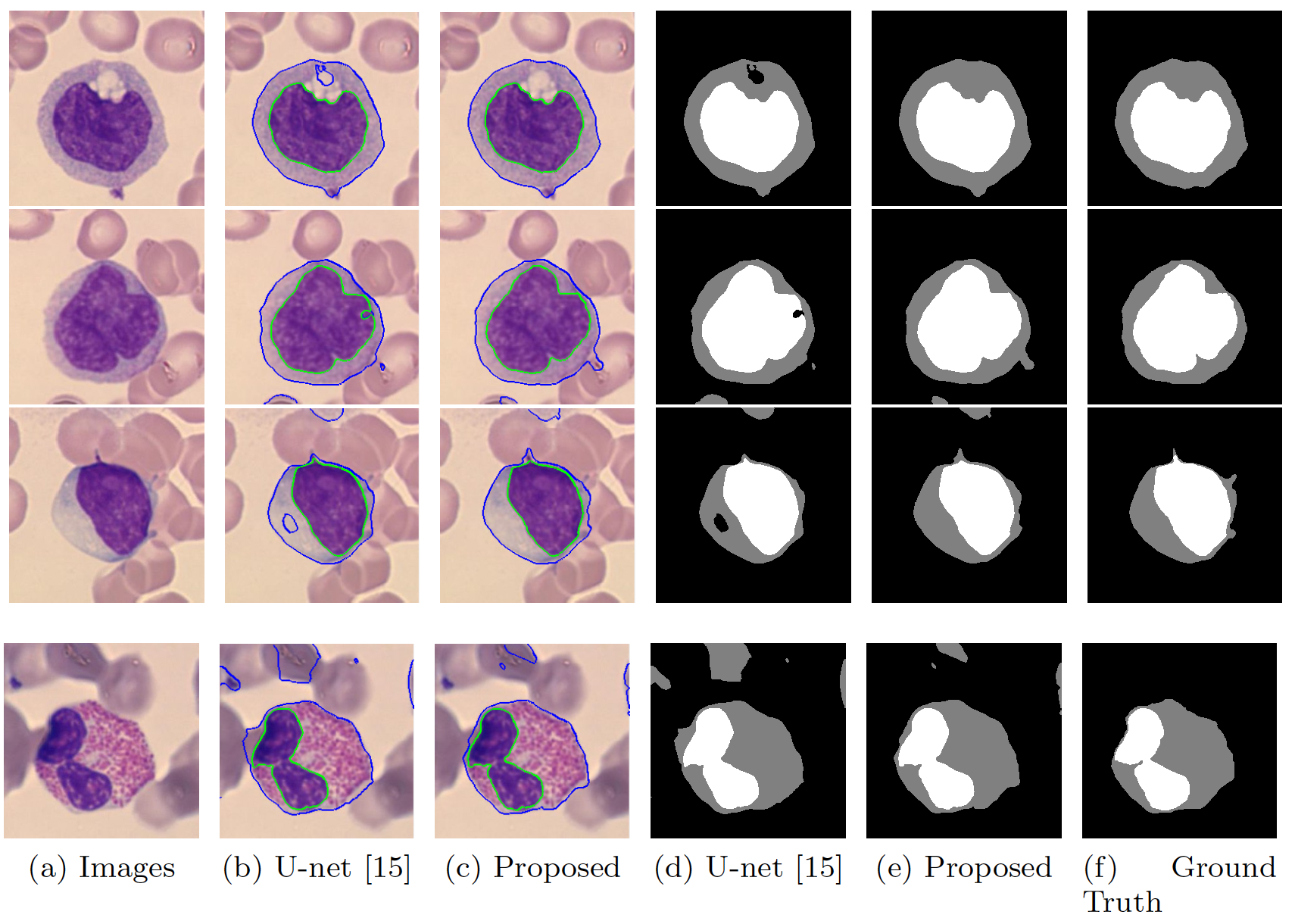}
\caption{Comparison of U-net\cite{Ronneberger2015} and the proposed VPTVU-net on WBC dataset.}\label{fig:wbc}
\end{figure}

The third experiment is to test the performance of the proposed VPTV-softmax on the popular U-net \cite{Ronneberger2015} in DCNN.  We adopt the basic structure of U-net \cite{Ronneberger2015} to get the similarity $\bm C$, denoted as $\bm o^K$, then the features $\bm o^K$ extracted by U-net are inputted into the proposed VPTV-softmax layer and get the dual variable $\bm q$ and $\bm f$ for spatial regularization and volume preserving. Finally, the summation of similarity $\bm o^K$, $\bm q$ and $\bm f$ is followed by a classical softmax activation function, the details of network architecture can be found in \figurename \ref{fig:net} and \ref{fig:dropout}. 
For convenience, we call the  modified U-net with the proposed VPTV-softmax as VPTVU-net. 
To use the classical softmax and for the algorithm stability, we let the entropic parameter $\varepsilon=1$ in this experiment.
We test U-net and our VPTVU-net on a White Blood Cell (WBC) dataset \cite{Zheng2018}. This dataset include $100$ WBC images with size $300\times 300$
which need to be segmented as $3$ phases: cell nucleus, WBC and background. In the background, there are many interferential red blood cells. We take $90$ images as the training sample and the rest $10$ images for test. In order to get more samples,
we randomly extracted $320\times 64=20480$ image patches with size $64\times 64$ from the original training images. In which, $85\%$ ($272\times 64$ patches) and $15\%$ ($48 \times 64$ patches) are taken as 
training and validation samples, respectively. 
For the volume constraints, we use the ratio of ground truth when the network is training.
For the test, the exact volume constraints are unavailable and we use the average ratios of training samples as an approximated volume constraints. This is just a very rough estimation. One flexibility of the proposed network is that the volume constraints $V$
can be set as a parameter according to different requirements when test the images.
The batch size of the training is set as $64$ and we use adam to optimize both of the networks with 20 epoches.
The training would be finished in half an hour on a linux server equipped with Maltab 2017a, a Tesla V100 GPU and matconvnet toolbox \cite{vedaldi15matconvnet}. When the training is finished, we take the test images with original size as the input of both of the networks, and get the final segmentation results.
The segmentation results and the related accuracy are listed in \figurename\ref{fig:wbc} and \tablename\ref{tab:1}.
In \figurename\ref{fig:wbc}, we list four test images for vision judgment, as can been seen from this figure, the proposed method 
can produce smoother segmentation boundaries and higher image accuracy than U-net. The dice distance is
used for the segmentation accuracy evaluation. Here this index is defined as 
$\frac{|\mathbb{L}\bigcap \mathbb{L}_{gt}|}{|\mathbb{L}|}\times 100 \%$, where $\mathbb{L}$ is segmented domain with different labels give by the algorithm and $\mathbb{L}_{gt}$ stands for the related ground truth domain. All the dice distances of the U-net and VPTVU-net for the 10 test images are listed in \tablename\ref{tab:1}. It implies the VPTV-softmax can get about $0.8\%$ improvements on U-net for this dataset.

\begin{table}[htp]
\caption{Accuracy of U-net and the proposed VPTVU-net for WBC dataset.}
\label{tab:1}
\begin{center}
\begin{tabular}{cccc}
\hline
 Images No.&U-net \cite{Ronneberger2015} & Proposed\\
\hline
\hline
1&97.27\% & 97.96\%\\
2&92.57\%& 96.81\%  \\
3&97.47\%& 97.95\%\\
4&97.66\%& 98.07\%\\
5&98.25\%& 98.89\%\\
6&98.46\%& 98.75\%\\
7&98.48\% & 99.12\%\\
8&98.29\%& 98.59\% \\
9&96.26\%& 96.67\%\\
10&95.89\%& 96.33\%\\          
\hline
\hline
Average&97.06\%&\textbf{97.91}\%\\
\hline
\end{tabular}
\end{center}
\end{table}

\begin{figure}
\centering
\includegraphics[width=1.0\linewidth]{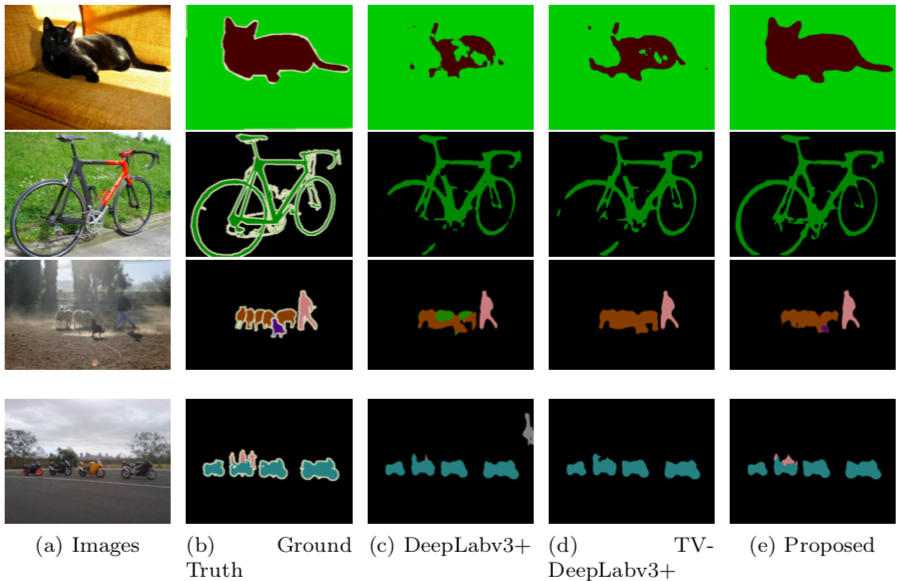}
\caption{Comparison of DeepLabv3+\cite{chen2018encoder}, TV-DeepLabv3+\cite{Jia2019} and the proposed VPTV-DeepLabv3+ on PASCAL-VOC2012 dataset.}\label{fig:DeepLabv3+1}
\end{figure}

\begin{figure}
\centering
\includegraphics[width=1.0\linewidth]{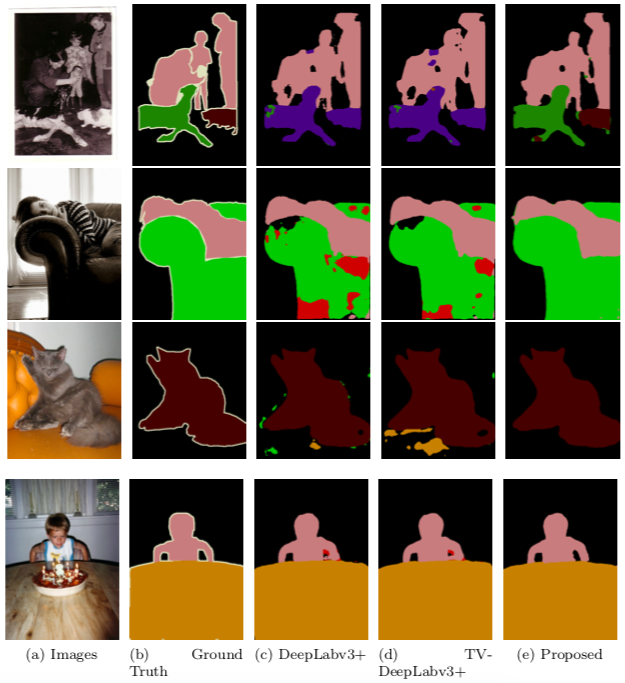}
\caption{The proposed VPTV-DeepLabv3+ will correct the wrong categories produced by the original DeepLabv3+\cite{chen2018encoder}.}
\label{fig:DeepLabv3+2}
\end{figure}

\begin{figure}
\centering
\includegraphics[width=1.0\linewidth]{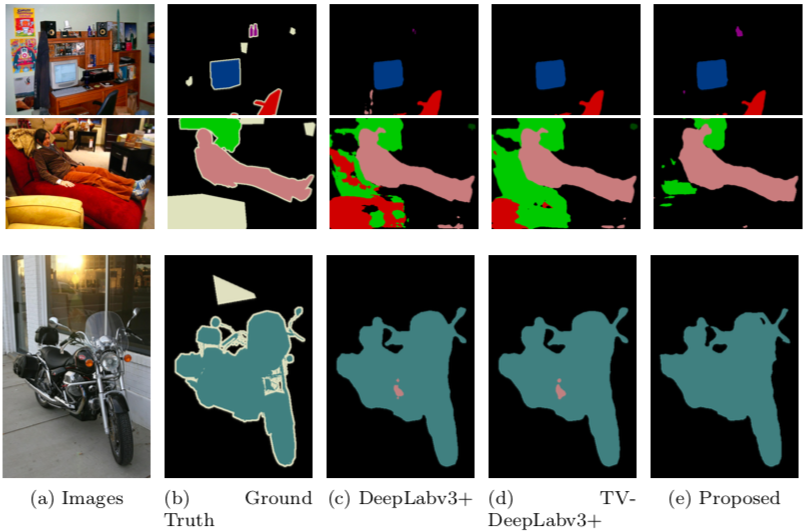}
\caption{Some classes with low probability may be lost in the segmentation results.}
\label{fig:DeepLabv3+3}
\end{figure}

The fourth experiment is to test the performance of the proposed VPTV-softmax on DeepLabv3+\cite{chen2018encoder} 
and compare the segmentation results produced by  DeepLabv3+, TV-softmax\cite{Jia2019} and our proposed VPTV-softmax. 
For convenience, we call the modified DeepLabv3+ with TV-softmax as TV-DeepLabv3+, and the modified DeepLabv3+ with the proposed VPTV-softmax as VPTV-DeepLabv3+. 
We test DeepLabv3+, TV-DeepLabv3+ and VPTV-DeepLabv3+ on the PASCAL-VOC2012 dataset. This dataset includes $20$ object classes and $1$ background class. We use the augmented training data ($10582$ images) obtained by standard preprocessing of $1464$ original images to train the network. The validation data includes $1448$ images to test the performance of the algorithms. In the numerical test, we let the entropic parameter $\varepsilon=0.5$, and the TV regularization parameter $\lambda=1.0$. Similar to DeepLabv3+, we use the xception-$65$ as the backbone network to extract features, other hyper-parameters are set
as the same as DeepLabv3+.  
For the volume preserving, the exact volume constraints are available from the ground truth when the network is training and test.  
The batch size of the training is set as $16$ and
the number of TV-softmax and VPTV-softmax layers
is $30$ (the iteration number $T$ in algorithm \ref{alg1}).
The segmentation results and the related accuracy are listed in \figurename \ref{fig:DeepLabv3+1}, \ref{fig:DeepLabv3+2}, \ref{fig:DeepLabv3+3} and \tablename~\ref{tab:3}.
In \figurename \ref{fig:DeepLabv3+1}, we list some test images for vision judgment. As can be seen from this figure, the TV-DeepLabv3+ can produce smoother segmentation boundaries than DeepLabv3+, and our proposed method not only can produce smoother segmentation boundaries than DeepLabv3+ but also higher image accuracy than TV-DeepLabv3+ due to the volume preserving. In the numerical experiment, we can also find that the segmentation results of our proposed method will correct the wrong categories produced by the original DeepLabv3+ in many cases, thus it can improve the segmentation accuracy. This fact is shown in \figurename \ref{fig:DeepLabv3+2}. However, as mentioned earlier, the entropic parameter $\varepsilon$ would affect the volume preserving, thus some classes with low probability may be lost in the segmentation result, which can be seen in \figurename \ref{fig:DeepLabv3+3}. Of course, there are also suboptimal segmentation results of the proposed method, which is inevitable. But the mean segmentation accuracy has improved a lot, as shown in \tablename\ref{tab:3}. 
Here the mean Intersection over Union(mIoU) is used for the segmentation accuracy evaluation. 
The mIoU of the DeepLabv3+, TV-DeepLabv3+ and VPTV-DeepLabv3+ with $30000$ and $50000$ iterations are listed in  \tablename\ref{tab:3}, respectively. 
Since we have set the batch size is $16$ (this can be done with 4 32G GPUs) and more iterations for training, the mIoU obtained here is higher than the report of the original DeepLabv3+ paper \cite{chen2018encoder} ($89\%$).
In this case, the VPTV-DeepLabv3+ still can get an average about $2\%$ improvements on DeepLabv3+ for this dataset.

\begin{table}[htp]
\caption{Accuracy of DeepLabv3+, TV-DeepLabv3+ and the proposed VPTV-DeepLabv3+ for PASCAL-VOC2012 dataset.}\label{tab:3}
\begin{center}
\begin{tabular}{c|c|c|c|c|c|c}
\hline
 Method&\multicolumn{2}{c|}{DeepLabv3+\cite{chen2018encoder}}&\multicolumn{2}{c|}{TV-DeepLabv3+\cite{Jia2019}}&\multicolumn{2}{c}{VPTV-DeepLabv3+}\\
\hline
\hline
Iteration number & 30000 & 50000 & 30000 & 50000 & 30000 & 50000\\
\hline
mIoU & 90.73\% & 91.45\% & 91.09\% & 91.71\% & \textbf{93.07}\% & \textbf{93.41}\%\\
\hline
\end{tabular}
\end{center}
\end{table}

\section{Conclusion}
In this paper, we have proposed a softmax segmentation with volume constraints. We show that 
the softmax in DCNN is a close-formed solution of an entropic regularization variational problem. With a volume preserving and the TV spatial 
regularization, we have given a variational formulation which is related to an entropic regularization optimal transport for image segmentation.
A dual algorithm has been proposed to solve the proposed model, and more importantly, this dual algorithm can be directly 
unrolled as a new VPTV-softmax layer for DCNN segmentation. The effectiveness of the proposed method is verified by the numerical experiments both on handcraft based model and deep learning based networks. 

Our idea in this paper is to replace the classical activation function in DCNN with a variational problem, this method implies that the priors such as convexity, connections, shapes in variational image segmentation can be easily extended to DCNN
based learning methods. We will have further research on these aspects.

\section{Appendix}
\subsection{Proof of Proposition \ref{pro2}}\label{proofpro2}
\textbf{Proof:} By introducing dual variables $\bm f,\bm g$ associated to the constraints $\mathrm{\boldsymbol{u}}\mathbf{1}_J=\bm V,\mathrm{\boldsymbol{u}}^\top\mathbf{1}_I=\mathbf{ 1}$, one can obtain
\begin{equation} 
\mathop{\mathrm{max}}\limits_{\bm f\in\mathbb{R}^I,\bm g\in\mathbb{R}^J}\langle\bm f,\bm V\rangle+\langle\bm g,\bm 1\rangle+\mathop{\mathrm{min}}\limits_{\boldsymbol{u}\geqslant0}\langle\mathrm{\mathbf{C}}+\mathbf{D}-\bm f\mathbf{1}_J^\top-\mathbf{1}_I\bm g^\top,\boldsymbol{u}\rangle-\varepsilon\mathrm{\mathbf{ H}}(\mathrm{\boldsymbol{u}}).
\label{Deduce1}
\end{equation} 
For $\varepsilon=0$, 
$$\mathop{\mathrm{min}}\limits_{\boldsymbol{u}\geqslant0}\langle\mathrm{\mathbf{C}}+\mathbf{D}-\bm f\mathbf{1}_J^\top-\mathbf{1}_I\bm g^\top,\boldsymbol{u}\rangle=
\begin{cases}
0, & \mathrm{if}~ \mathrm{\mathbf{C}}+\mathbf{D}-\bm f\mathbf{1}_J^\top-\mathbf{1}_I\bm g^\top\geqslant0 \\
-\infty, & \mathrm{otherwise}
\end{cases}$$
so that the constraint reads $\mathrm{\mathbf{C}}+\mathbf{D}-\bm f\mathbf{1}_J^\top-\mathbf{1}_I\bm g^\top=\mathrm{\mathbf{C}}+\mathbf{D}-\bm f\oplus\bm g\geqslant0$.\\
For $\varepsilon>0$, by standard discussion in the latter optimal problem, one has
\begin{equation}  
u_{i,j}^{*}=e^{\frac{-\mathbf{C}_{i,j}-\mathbf{D}_{i,j}+f_i^{*}+g_j^{*}}{\varepsilon}-1}.
\end{equation}
Substitute $\boldsymbol{u}_{i,j}^{*}$ into the formula \eqref{Deduce1}, one can obtain \eqref{DualProblem1}. \hfill $\blacksquare$

\renewcommand{\refname}{References}
\bibliographystyle{unsrt}
\bibliography{reference}


\end{document}